# Some Properties of Joint Probability Distributions

**Marek J. Druzdzel**
University of Pittsburgh
Department of Information Science
Pittsburgh, PA 15260
*marek@lis.pitt.edu*

## Abstract

Several Artificial Intelligence schemes for reasoning under uncertainty explore either explicitly or implicitly asymmetries among probabilities of various states of their uncertain domain models. Even though the correct working of these schemes is practically contingent upon the existence of a small number of probable states, no formal justification has been proposed of why this should be the case. This paper attempts to fill this apparent gap by studying asymmetries among probabilities of various states of uncertain models. By rewriting the joint probability distribution over a model's variables into a product of individual variables' prior and conditional probability distributions and applying central limit theorem to this product, we can demonstrate that the probabilities of individual states of the model can be expected to be drawn from highly skewed lognormal distributions. With sufficient asymmetry in individual prior and conditional probability distributions, a small fraction of states can be expected to cover a large portion of the total probability space with the remaining states having practically negligible probability. Theoretical discussion is supplemented by simulation results and an illustrative real-world example.

## 1 INTRODUCTION

One way of looking at models of uncertain domains is that those models describe a set of possible states of the world,[1] only one of which is true. This view is explicated by the logical Artificial Intelligence (AI) approaches to reasoning under uncertainty — at any given point various extensions of the current body of facts are possible, one of which, although unidentified, is assumed to be true. The number of possible extension of the facts is exponential in the number of uncertain variables in the model. It seems to be intuitively appealing, and for sufficiently large domains practically necessary, to limit the number of extensions considered. Several AI schemes for reasoning under uncertainty, such as case-based or script-based reasoning, abduction (e.g., (Charniak & Shimony, 1994)), non-monotonic logics, as well as recently proposed search-based methods for belief updating in Bayesian belief networks (e.g., (Henrion, 1991; Poole, 1993)), seem to be following this path. If a domain is uncertain and any of the exponential number of extensions of observations is possible, one might ask why concentrating merely on a small number of them would work. In this paper, I show that we can usually expect in uncertain models a small fraction of all possible states to account for most of the total probability space.

My argument refers to models rather than the systems that they describe. As argued elsewhere (Druzdzel & Simon, 1993), deriving results concerning models and relating these results to reality is the best that we can hope for as scientists. I will assume that for any uncertain domain, there exists a probabilistic model of that domain, even though in some cases construction of a probabilistic model may be impractical. In all derivations and proofs, for the reasons of convenience, I will consider only discrete probability distributions. The analysis can be generalized to continuous distributions by, for example, discretizing them or considering intervals over probabilities (in particular, infinitesimally small intervals). Further, I will be making certain assumptions when applying central limit theorem. I believe these assumptions to be sufficiently weak to refer to the applicability of the argument as "most of the time" or "usually the case."

The remainder of this paper is structured as follows. Section 2 describes the probabilistic framework for representing uncertainty, outlines my approach to studying the properties of joint probability distributions over models, and briefly discusses applicability of central limit theorem to this analysis. Section 3 presents the main argument of the paper. Section 3.1 discusses the general case, where conditional probability distributions are arbitrary. In this case, any probability within the joint probability distribution can be expected to come from a lognormal distribution, although each probability can be possibly drawn from a distribution with different parameters. It is shown that if the individual conditional probabilities are sufficiently extreme, then a small fraction of the most likely states will cover most of the probability space. Section 3.2 looks at the simplest special case, where each of the conditional distributions of the model's variables is identical, showing that probabilities within the joint probability distribution are distributed lognormally. Section 3.3 extends this result to the case where conditional distributions are not identical, but are identically distributed. Section 3.4 argues that there are good reasons to expect that the special case result may be a good approximation for most practical models. Section 4 analyzes the joint probability distribution of ALARM, a probabilistic model for monitoring anesthesia patients, showing empirical support for the earlier theoretical derivations. Finally, Section 5 discusses the implications of the discussed properties of uncertain domains for uncertain reasoning schemes.

---

[1] A state can be succinctly defined as an element of the Cartesian product of sets of outcomes of all individual model's variables. There is a richness of terms used to describe states of a model: extension, instantiation, possible world, scenario, etc. Throughout this paper, I will attempt to use the term *state of a model* or briefly *state* whenever possible.

## 2 PRELIMINARIES

### 2.1 PROBABILISTIC MODELS

The essence of any probabilistic model is a specification of the joint probability distribution over the model's variables, i.e., probability distribution over all possible deterministic states of the model. It is sufficient for deriving all prior, conditional, and marginal probabilities of the model's individual variables[2].

Most modern textbooks on probability theory relate the joint probability distribution to the interactions among variables in a model by factorizing it, i.e., breaking it into a product of priors and conditionals. While this view has its merits in formal expositions, it suggests viewing a probabilistic model as merely a numerical specification of a joint probability distribution that can be possibly algebraically decomposed into factors. This clashes with our intuition that whatever probability distributions we observe, they are a product of structural, causal properties of the domain. Causal interactions among variables in a system determine the observed probabilistic dependences and, in effect, the joint probability distribution over all model's variables. An alternative view of a joint probability distribution is, therefore, that it is *composable from* rather than *decomposable into* prior and conditional probability distributions. In this view, each of these distributions corresponds to a causal mechanism acting in the system (Druzdzel & Simon, 1993). This reflects the process of constructing joint probability distributions over domain models in most practical situations.

Since insight obtained from two modeling tools: Bayesian belief networks (BBNs) (Pearl, 1988) and probability trees may prove useful for the reader, I will show how they both represent a simple uncertain model involving a common activity of a clinician interpreting the result of a screening test for a disease. This model contains two binary variables: *disease* and *test*. The outcomes of variable *disease*, $d$ and $\bar{d}$, stand for *disease present* and *disease absent* respectively. The outcomes of variable *test*, $t$ and $\bar{t}$, stand for *test positive* and *test negative* respectively. A BBN representing this problem, shown in Figure 1, reflects the qualitative structure of the domain, showing explicitly dependences among variables. Each variable is characterized by a probability distribution conditional on its predecessors or by a prior probability distribution if the variable has no predecessors. Figure 1 shows also a probability tree encoding the same problem. Each node in this tree represents a random variable and each branch originating from that node a possible outcome of that variable. Each complete path starting at the root of the tree and ending at a leaf corresponds to one of the four possible deterministic states of the model.

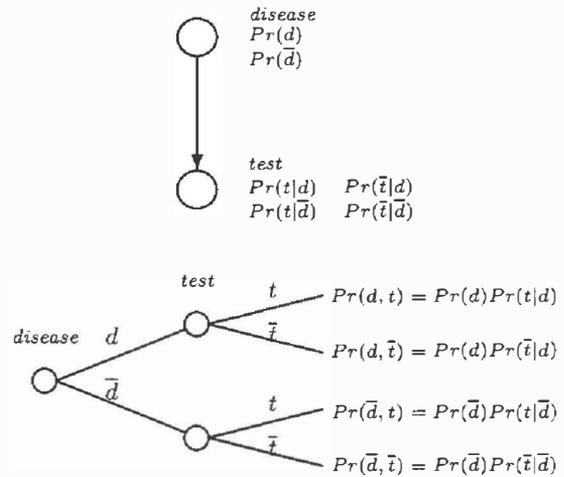

Figure 1: Two probabilistic representations of the screening test problem: Bayesian belief network (upper) and probability tree (lower).

The probabilities of various states of a model can be easily retrieved in BBNs and probability trees by multiplying out the prior and conditional probabilities of individual variables. In the models of Figure 1, we multiply the priors of various outcomes of *disease* by the conditionals of respective outcome of *test* given presence or absence of *disease*.

---

[2]I will often refer to the prior probability distribution over a variable as "prior" and a conditional probability distribution over a variable's outcomes given the values of other model's variables as "conditional."

## 2.2 STATE PROBABILITIES

Let us choose at random one state of a model that consists of $n$ variables $X_1, X_2, X_3, \ldots, X_n$. We choose this state equiprobably from among all possible states, regardless of its probability. One way of imagining this is that we are drawing a marble out of a basket containing uniquely marked but otherwise identical marbles. As a state is an instantiation of each of the model's $n$ variables, another way of looking at this selection process is that we are traversing the probability tree representing the model from its root to one of its leaves taking at each step one of the possible branches with equal probability. This amounts to a random choice of one outcome from among the outcomes of each of the variables. For example, we might randomly select one of the four states in the model of Figure 1 by first choosing one of the two possible outcomes of the variable *disease* by flipping a coin (let the outcome be for example $d$) and then choosing one of the two possible outcomes of the variable *test* by flipping a coin again (let this outcome be for example $t$). Our procedure made selection of each state equiprobable (with probability 0.25 in our example). The probability $p$ of a selected state is equal to the product of conditionals of each of the randomly selected outcomes. It is equal for our selected state to $p = Pr(d,t) = Pr(d)Pr(t|d)$. In general, if we denote $p_i$ to be the conditional (or prior) probability of the randomly selected outcome of variable $X_i$, we have

$$p = p_1 p_2 p_3 \ldots p_n = \prod_{i=1}^{n} p_i \ . \quad (1)$$

In random selection of a state, we chose each $p_i$ to be one number from among the probabilities of various outcomes of variable $X_i$. We can, therefore, view each $p_i$ as a random variable taking equiprobable values from among the probabilities of the outcomes of variable $X_i$. Of course, the distribution of $p_i$ is not in general independent from the distribution of $p_j$, $i \neq j$, as the outcomes of some variables may impact the conditional probability distributions of other variables. Selection of $p_i$ within its distribution, however, is independent of any other $p_j$, $i \neq j$. Note that in our simple example we used outcomes of independent coin tosses to choose a state. Intuitively, if the model is causal, then even though the mode in which a mechanism is working, described by a conditional probability distribution, depends on the outcomes of its causal ancestors, the exact form of this distribution (i.e., the values of probabilities of different outcomes) is a property of the mechanism and is independent on anything else in the system.

Having described the process of randomly drawing a state as above, can we say anything meaningful about the distribution of $p$? It turns out that we can say quite a lot about a simple transformation of $p$. By taking the logarithm of both sides of (1), we obtain

$$\ln p = \ln \prod_{i=1}^{n} p_i = \sum_{i=1}^{n} \ln p_i \ . \quad (2)$$

As for each $i$, $p_i$ is a random variable, its logarithm $\ln p_i$ is also a random variable, albeit with a different distribution. The asymptotic behavior of a sum of random variables is relatively well understood and addressed by a class of limit theorems known collectively as central limit theorem. When the number of components of the sum approaches infinity, the distribution of the sum approaches normal distribution, regardless of the probability distributions of the individual components. Even though in any practical case we will be dealing with a finite number of variables, the theorem gives a good approximation even when the number of variables is small.

## 2.3 CENTRAL LIMIT THEOREM: "ORDER OUT OF CHAOS"

Central limit theorem (CLT) is one of the fundamental and most robust theorems of statistics, applicable to a wide range of distributions. It was originally proposed for Bernoulli variables, then generalized to independent identically distributed variables, then to non-identically distributed, and to some cases where independence is violated. Extending the boundaries of distributions to which CLT is applicable is one of active areas of research in statistics. CLT is so robust and surprising that it is sometimes referred to as "order out of chaos" (de Finetti, 1974).

One of the most general forms of CLT is due to Liapounov (to be found in most statistics textbooks).

**Theorem 1** *Let $X_1, X_2, X_3, \ldots, X_n$ be a sequence of $n$ independent random variables such that $E(X_i) = \mu_i$, $E((X_i - \mu_i)^2) = \sigma_i^2$, and $E(|X_i - \mu_i|^3) = \omega_i^3$ all exist for every $i$. Then their sum, $Y = \sum_{i=1}^{n} X_i$ is asymptotically distributed as $N(\sum_{i=1}^{n} \mu_i, \sum_{i=1}^{n} \sigma_i^2)$, provided that*

$$\lim_{n \to \infty} \frac{\sum_{i=1}^{n} \omega_i^3}{\left(\sum_{i=1}^{n} \sigma_i^2\right)^{3/2}} = 0 \ . \quad (3)$$

If the variables $X_i$ are identically distributed, i.e., when $\forall_{1 \leq i \leq n} \ \mu_i = \mu$, $\sigma_i = \sigma$, and $\omega_i = \omega$, (3) reduces to

$$\lim_{n \to \infty} \frac{n \omega^3}{n^{3/2} \sigma^3} = \lim_{n \to \infty} \frac{\omega^3}{\sqrt{n} \sigma^3} = 0 \ .$$

This condition is satisfied for any distribution for which $\mu$ and $\sigma$ exist and the theorem reduces to Lindeberg and Lévy's version of CLT (also reported in most textbooks).

Returning to Equation (2), we have by the CLT, that assuming that the preconditions of CLT are satisfied, the sum on the right side is in the limit normally distributed. If $\ln p$ is normally distributed, then $p$ itself must be drawn from a lognormal distribution.

## 3 PROPERTIES OF THE JOINT PROBABILITY DISTRIBUTION

CLT captures the growth of a process showing strong regularity and satisfying certain independence conditions. For the purpose of this paper, I choose to demonstrate that these conditions are reasonably satisfied in the process of constructing a joint probability distribution. I will argue that the type of process that we are dealing with is one that is addressed by the theorem.

In what follows, I will be studying the properties of the logarithm of the distribution rather than the distribution itself. This is motivated by a practical consideration — the lognormal distributions resulting from the application of the CLT tend to span over many orders of magnitude and are extremely skewed. Logarithmic scale can be most appreciated in the diagrams that will be shown later in the paper — the skewness of the distributions makes diagrams drawn in linear scale practically unreadable. Changing the scale between logarithmic and linear is straightforward.

### 3.1 ARBITRARY CONDITIONAL PROBABILITY DISTRIBUTIONS

#### 3.1.1 Preconditions

Establishing the circumstances under which the condition (3) of CLT holds for multiply valued variables is not straightforward. It turns out that the condition is satisfied for propositional variables under a weak assumption, requiring only that the sequence of variances of the variables in the model is divergent in the limit.

Let a variable $X_i$ have two outcomes: $x_i$ with probability $p_i$ and outcome $\overline{x_i}$ with probability $1 - p_i$. The mean $\mu_i$, variance $\sigma_i^2$, and $\omega_i$ over the logarithm of this probability distribution are

$$\mu_i = \frac{1}{2}\ln(p_i(1-p_i))$$
$$\sigma_i^2 = \frac{1}{4}\ln^2 \frac{p_i}{1-p_i}$$
$$\omega_i^3 = \frac{1}{8}\left|\ln \frac{p_i}{1-p_i}\right|^3$$

It is clear, that for any variable $i$, $|\sigma_i| = |\omega_i|$. Equation (3) transforms into

$$\lim_{n\to\infty} \frac{\sum_{i=1}^n \omega_i^3}{\left(\sum_{i=1}^n \sigma_i^2\right)^{3/2}} = \lim_{n\to\infty} \frac{\sum_{i=1}^n \omega_i^3}{\left(\sum_{i=1}^n \sigma_i^2\right)\sqrt{\sum_{i=1}^n \sigma_i^2}}$$
$$= \lim_{n\to\infty} \frac{\sum_{i=1}^n \omega_i^3}{\sum_{i=1}^n |\sigma_i|^3 + \sum_{i=1}^n \Delta_i \sigma_i^2} = 0 \ .$$

where $\Delta_i > 0$ for $1 \leq i \leq n$. If the sequence of variances is divergent, the denominator becomes in the limit infinitely larger than the numerator and the whole expression will approach zero, as required by (3).

#### 3.1.2 Distribution of Probabilities of States

Let a model consist of $n$ variables $X_1, X_2, X_3, \ldots, X_n$, having $k_1, k_2, k_3, \ldots, k_n$ outcomes respectively ($1 \leq i \leq n$). For any single state, we can apply the CLT to (2), viewing each $p_i$ as an independent random variable. The value of $p_i$ will be the probability of a randomly selected outcome of variable $X_i$ (equiprobably selected value from among the probability distribution of $X_i$). Let the mean and the variance of the distribution of $p_i$ be $\mu_i$ and $\sigma_i^2$ respectively. The logarithm of $p$, the probability of an individual state, obtained by multiplying priors and conditionals of individual variables is then distributed as $\ln p \sim N(\sum_{i=1}^n \mu_i, \sum_{i=1}^n \sigma_i^2)$. The density function $f(\ln p)$ is

$$f(\ln p) = \frac{1}{\sqrt{2\pi \sum_{i=1}^n \sigma_i^2}} \exp \frac{-(\ln p - \sum_{i=1}^n \mu_i)^2}{2\sum_{i=1}^n \sigma_i^2} \ ,$$

where $0 < p < 1$.

#### 3.1.3 Expected Probability Mass of Various States

While $f(\ln p)$ expresses the distribution of the logarithm of the probability of the randomly chosen state, a normalized $pf(\ln p)$ will express the expected contribution of all states with probability $p$ to the total probability mass (in logarithmic scale).

$$pf(\ln p) = \frac{p}{\sqrt{2\pi \sum_{i=1}^n \sigma_i^2}} \exp \frac{-(\ln p - \sum_{i=1}^n \mu_i)^2}{2\sum_{i=1}^n \sigma_i^2}$$
$$= \frac{1}{\sqrt{2\pi \sum_{i=1}^n \sigma_i^2}}$$
$$\quad \exp \frac{-(\ln p - \sum_{i=1}^n \mu_i)^2 + 2\sum_{i=1}^n \sigma_i^2 \ln p}{2\sum_{i=1}^n \sigma_i^2}$$
$$= \frac{1}{\sqrt{2\pi \sum_{i=1}^n \sigma_i^2}} \exp \frac{\sum_{i=1}^n \sigma_i^2 + 2\sum_{i=1}^n \mu_i}{2}$$
$$\quad \exp \frac{-(\ln p - (\sum_{i=1}^n \mu_i + \sum_{i=1}^n \sigma_i^2))^2}{2\sum_{i=1}^n \sigma_i^2}$$
$$= C \exp \frac{-(\ln p - (\sum_{i=1}^n \mu_i + \sum_{i=1}^n \sigma_i^2))^2}{2\sum_{i=1}^n \sigma_i^2} \quad (4)$$

This function belongs the same class as $f(\ln p)$. $C$ is a normalizing constant that makes the integral of the function for $-\infty < \ln p < 0$ equal to 1.0.

Several important qualitative properties of this function are determined purely by its form. As $\sum_{i=1}^n \sigma_i^2$ is positive, the function is shifted with respect to $f(\ln p)$ by $\sum_{i=1}^n \sigma_i^2$ towards higher values of probabilities. This shift depends on both the number of variables in the model and the individual variances in probabilities: it is stronger for distributions with high variance (i.e., distributions that show stronger asymmetries in probabilities of various outcomes). The contribution function reaches its maximum for higher probabilities than the distribution of probabilities of states. (Exactly for $\ln p = \sum_{i=1}^n \mu_i + \sum_{i=1}^n \sigma_i^2$ or for $\ln p = 0$, whichever is lower. The zero cutoff point reflects the fact that the distribution is defined only for $-\infty < \ln p < 0$.) For $0 \leq \sum_{i=1}^n \mu_i + \sum_{i=1}^n \sigma_i^2$, we will observe the pattern that a few most likely states explain most of the probability mass. Large variance in individual distributions makes, therefore, not only the distribution of probabilities spread over many orders of magnitude but accounts for the shift of the distribution of their contributions to the total probability mass larger. The more we know about a domain, the more asymmetry individual conditionals will show. When the domain and its mechanisms are well known, probability distributions tend to be extreme. This implies a large variance and a large shift in the expected contribution function and, therefore, a small number of very likely states of the model. This makes intuitive sense — we tend to act with confidence in environments that we know well, just because we can easily predict what will happen. When an environment is less familiar, the probability distributions tend to be less extreme, there is less variance in probabilities. The shift in contribution function is small and none of the states is very likely.

### 3.2 IDENTICAL CONDITIONAL PROBABILITY DISTRIBUTIONS

The previous section has derived limiting distribution for each individual state. Although this gives much insight into expected asymmetries in uncertain domains, of more practical interest is the distribution over all probabilities of the joint probability distribution. Clearly, there are cases for which we will be able to derive these. One such special case is when the sums of means and variances of individual distributions are always the same for each state. This may be, for example, the case when all variables are independent or when all conditional probability distributions of each of the model's variables are permutations of one another. Since we are not concerned with the outcomes of variable $X_i$ but their probabilities, it does not matter how these probabilities are assigned to the outcomes. This special case is the subject of the current section. The next section will relax this to the case where the conditionals for each of the variables have the same means and variances.

Let each of the random variables $X_i$ have $k$ outcomes and let each probability $p_i$ in (1) come from the same probability distribution (i.e., each variable's conditionals are permutations of some discrete probability distribution over $k$ outcomes). Since in generating the probability of a randomly selected state we choose one of the possible $k$ outcomes of each variable equiprobably, we have the mean and the variance of the distribution of each individual variable $p_i$ in the product equal to

$$\mu = \frac{1}{k}\sum_{i=1}^{k} \ln p_i, \quad \sigma^2 = \frac{1}{k}\sum_{i=1}^{k} (\ln p_i - \mu)^2 .$$

Since the distribution of each of the factors is the same, it is independent of other factors. Given independence of distributions and the fact that both the mean $\mu$ and the variance $\sigma^2$ exist, we can apply Lindeberg and Lévy's version of the CLT to (2), obtaining $\ln p \sim N(n\mu, n\sigma^2)$. The density function $f(\ln p)$ is

$$f(\ln p) = \frac{1}{\sigma\sqrt{2\pi n}} \exp \frac{-(\ln p - n\mu)^2}{2n\sigma^2}, \quad 0 < p < 1 .$$

The normalized $pf(\ln p)$ of Equation (4), expressing the expected contribution of all states with probability $p$ to the total probability mass becomes

$$pf(\ln p) = C \exp \frac{-(\ln p - n(\mu + \sigma^2))^2}{2n\sigma^2} .$$

As shown in Section 3.1, this function will be shifted towards higher values of probabilities, as $n\sigma^2$ is positive. The magnitude of this shift depends on the variances in individual probability distributions.

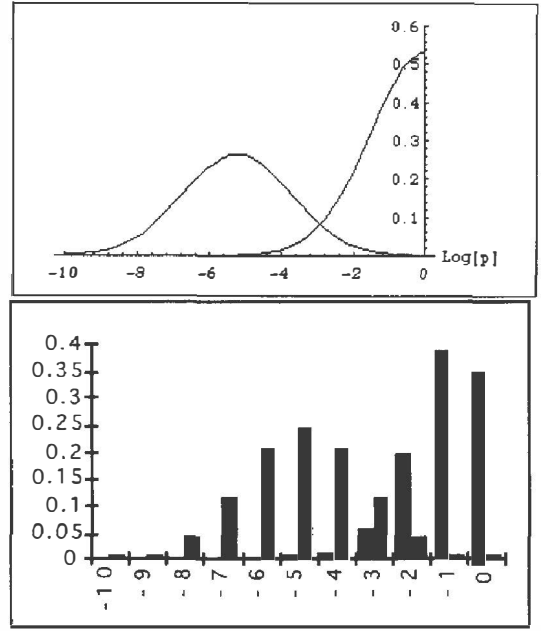

Figure 2: Identical conditional probability distributions for 10 binary variables with probabilities of outcomes equal to 0.1 and 0.9: theoretically derived probability distribution over probabilities of states of the joint probability distribution and the distribution of their contribution to the probability mass (upper diagram) and the histograms observed in a simulation (lower diagram).

The upper diagram in Figure 2 shows this theoretically derived relationship for a model consisting of $n = 10$ binary ($k = 2$) variables with probability distributions $p_1 = 0.1$ and $p_2 = 0.9$.[3] Please, note that the distribution of the contributions of probabilities of states to the total probability mass is strongly shifted towards higher probabilities and cut off at point $\log p = 0$. The lower diagram in Figure 2 shows the result of a simulation in which an uncertain model satisfying the assumption was randomly created and then its joint probability distribution analyzed. This simulation was done in the spirit of a demonstration device similar to those proposed by Gauss or Kapteyn to show a mechanism by which a distribution is generated. Similarity of the theoretically derived distributions to the simulation results, even for as few as 10 random variables, is apparent. Figure 3 shows theoretically derived distribution functions for similar models, in which individual probability distributions were $p_1 = 0.2$ and $p_2 = 0.8$ (upper diagram) and $p_1 = 0.3$ and $p_2 = 0.7$ (lower diagram). With smaller variances in probabilities (the distributions are closer to being symmetric), the shift is much smaller. In such cases, most states will have low probabilities and, hence, no very likely states will be observed.

---

[3]This and other figures use decimal rather than natural logarithm because of the ease with which we can translate the value of the decimal logarithm to order of magnitude in the decimal system.

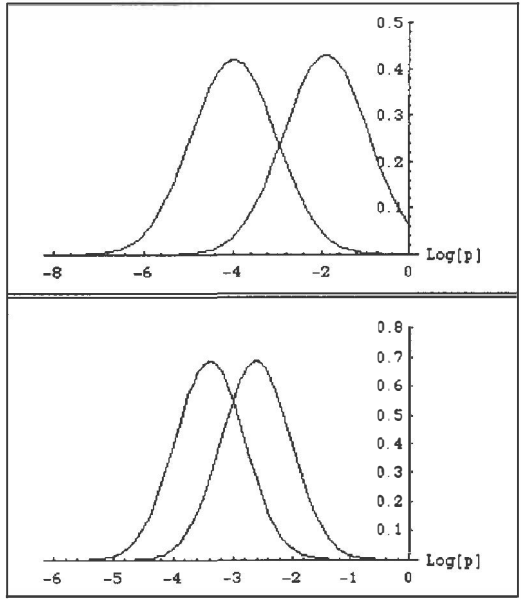

Figure 3: Theoretically derived distributions for identical conditional probability distributions for 10 binary variables with probabilities of outcomes equal to 0.2 and 0.8 (upper diagram) and 0.3 and 0.7 (lower diagram).

Since $\ln p$ follows normal distribution, $p$ will be lognormally distributed. This distributions will usually be highly positively skewed, even for such moderate values of probabilities as 0.1. The skewness coefficient of the distribution of Figure 2, for example, is $\gamma \approx 7.3 \times 10^7$.

Fitting a lognormal distribution with parameters $\xi$ and $\phi$ to the distribution of probabilities within joint probability distribution allows for determining the probability threshold $t$ for which all states less likely than $t$ contribute totally less than fraction $f$ of the total probability space. This threshold $t$ can be used as an error bound in search-based belief updating algorithms. It is the solution to the following equation

$$\int_{-\infty}^{t} p f(\ln p) d\ln p$$
$$= \int_{-\infty}^{t} C \exp \frac{-(\ln p - (\xi + \phi^2))^2}{2\phi^2} d\ln p = f .$$

This equation does not have a closed-form solution. After solving it numerically, we can easily convert $t$ into the fraction of states $l$ that are less likely than $t$.

$$l = \int_{-\infty}^{f} f(\ln p) d\ln p$$
$$= \int_{-\infty}^{f} \frac{1}{\phi\sqrt{2\pi}} \exp \frac{-(\ln p - \xi)^2}{2\phi^2} d\ln p .$$

### 3.3 IDENTICALLY DISTRIBUTED CONDITIONAL PROBABILITY DISTRIBUTIONS

Suppose that instead of identical probabilities, we have probabilities coming from the same probability distribution. Let the mean and the variance of this distribution be $\mu$ and $\sigma^2$ respectively. We are still able to apply Lindeberg and Lévy's version of the theorem to (2) as the distribution of each factor $\ln p_i$ is independent of the distribution of any other factor. Figure 4 shows the analogue of Figure 2 for both the theoretically derived relationship and simulation results, where the model consists of $n = 10$ binary ($k = 2$) variables with probability distributions drawn uniformly from the intervals $[0, 0.1]$ and $[0.9, 1.0]$.

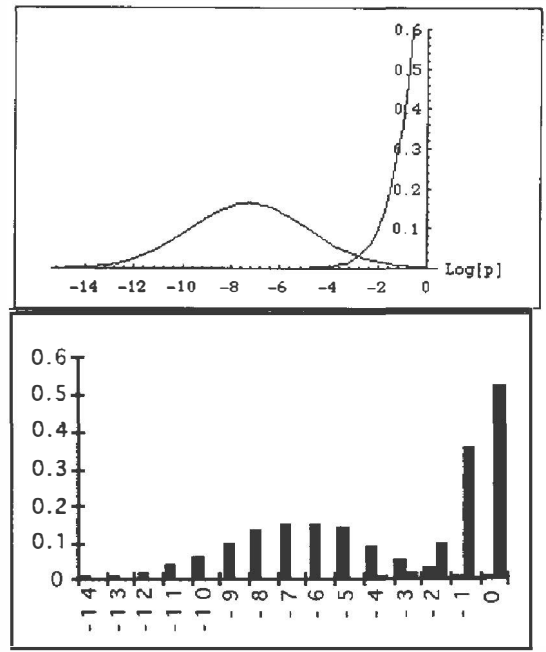

Figure 4: Identically distributed conditional probability distributions for 10 binary variables with probabilities of outcomes drawn uniformly from the intervals $[0, 0.1]$ and $[0.9, 1.0]$: theoretically derived probability distribution over probabilities of states of the joint probability distribution and the distribution of their contribution to the probability mass (upper diagram) and the histograms observed in a simulation (lower diagram).

## 3.4 EXPECTATIONS REGARDING PRACTICAL MODELS

The general case result of Section 3.1, showing that each probability in the joint probability distribution comes from a lognormal distribution, although each with perhaps different parameters, is rather conservative. In fact, CLT is known for its robustness and violations of the preconditions for the theorem may simply affect the speed of convergence rather than the normality of the sum. There are several intuitive reasons for why the distribution over probabilities of different states of a model might approach the lognormal distribution in most practical models. Conditional probabilities in practical models tend to belong to modal ranges, at most a few places after the decimal point, such as between 0.001 and 1.0. This may be an artifact caused by experts' tendency to use landmark probabilities that not only make various distributions modal, but also similar to one another. Another reason for this is that interactions characterized by extremely small probabilities may be excluded from models as not relevant. Translated into the decimal logarithmic scale, it means the interval between −3 and 0, which is further averaged over all probabilities (that have to add up to one) and for variables with few outcomes will result in mean probabilities that belong to even more modal ranges. In effect, even though each probability in the joint probability distribution comes from a different lognormal distribution, the parameters of these distributions may be quite close to one another. $\sum \mu$ and $\sum \sigma^2$ are unlikely to show large variation and there will be many similar values. Topology of the model, i.e., the connectivity of the underlying graph can be expected to have influence on the goodness of fit — as sparsely connected graphs contain less dependences, they should provide a better fit. Most practical graphs seem to be sparsely connected. Finally, the limit effects expressed by the CLT may be robust against dependences between conditional distributions of various variables. It is not unreasonable to expect that in many practical models, the distribution of probabilities of the model states the joint probability distributon will approach lognormality.

## 4 EXAMPLE: ALARM

Given the strength of the conclusions of the theoretical analysis, it might be useful to study the properties of the joint probability distribution over a real model. The most realistic model with a full numerical specification that was available to me was ALARM, a medical diagnostic model of monitoring anesthesia patients in intensive care units (Beinlich et al., 1989). With its 38 random variables, each having two or three outcomes, ALARM has a computationally prohibitive number of states. I selected, therefore, several self-contained subsets of ALARM consisting of 7 to 13 variables, and analyzed the distribution of probabilities of all states within those subsets.

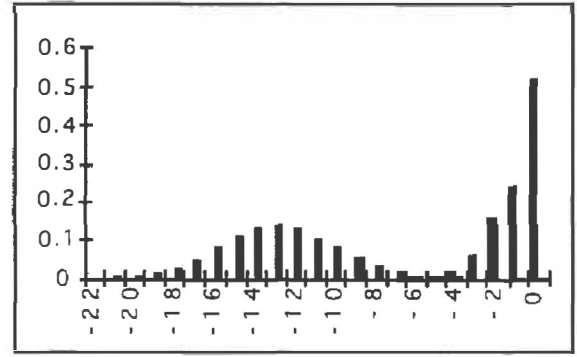

Figure 5: Histograms of the probabilities of various states (the bell-shaped curve) and their contribution to the total probability space (the peak on the right side) for a subset of 13 variables in ALARM model.

Figure 5 shows the result of one of such run, identical with the results of all other runs with respect to the form of the observed distribution. It is apparent that the histogram of states appears to be for normally distributed variables, which, given that the ordinate is in logarithmic scale, supports the theoretically expected lognormality of the actual distribution. The histogram also indicates and small contribution of its tail to the total probability mass. The subset studied contained 13 variables, resulting in 525,312 states. The probabilities of these states were spread over 22 orders of magnitude. Of all states, there was one state with probability 0.52, 10 states with probabilities in the range (0.01, 0.1) and the total probability of 0.23, and 48 states with probabilities in the range (0.001, 0.01) and the total probability of 0.16. The most likely state covered 0.52 of the total probability space, the 11 most likely states covered 0.75 of the total probability space, and the 49 most likely states (out of the total of 525,312) covered 0.91 of the total probability space.

## 5 CONCLUSION

Using a hypothetical probabilistic model of a typical uncertain domain, I have demonstrated that the joint probability distribution over its variables is created by a multiplicative process, combining conditional probabilities of individual variables. Asymmetries in these individual distributions, which I argued can be expected because of structural properties of models, result in joint probability distributions exhibiting orders of magnitude differences in probabilities of various states of the model. In particular, there is usually a small fraction of states that cover a large portion of the total probability space with the remaining states having practically negligible probability.

Even though I referred to models as wholes, the asymmetries derived in the preceding sections will hold for their self-contained parts. Having a large probabilistic model, we can at each reasoning step determine what, if anything, is relevant for a given query.[4] If this selected part contains random variables, it is amenable to our argument.

The analysis contained in this paper concerns static systems. I believe, however, that the essential argument is easily transferable into dynamic systems. Note, that to model a transition of a system over time, we can replace each variable $X$ in a static model by additional variables $X_{t_i}$ specifying the state of $X$ at time $t_i$. The value of variable $X_{t_i}$ can be specified by a probability distribution conditional on the values of variables $X_{t_j}, j < i$. Then, it is straightforward to extend the argument to all newly introduced variables in the way I did for static systems.

Two special classes of interest are variables with symmetric probability distributions, such as the outcomes of die tosses, and deterministic variables. If all of the variables in a model had symmetric distributions or all of them were deterministic, then the denominator of (3) would be zero. Clearly, the argument will hold as long as there is a sufficient number of variables that belong to the complement of these two classes. Variables with symmetric distributions simply tend to shift the distribution of the probability mass towards lower values, decreasing the expected contributions of the most likely states, while deterministic variables achieve the opposite.

The significance of this analysis is that it provides a clarification for what has been long assumed but never, to my knowledge, explicated. By providing a framework for studying the distribution of probabilities of individual states in the joint probability distribution, this analysis provides foundations for one direction of research on approximate reasoning schemes that are correct and yet computationally tractable. The observed and theoretically derived asymmetry in the distribution of probabilities of individual states of the model suggests that considering only a small number of them can lead to good approximations in belief updating. One possible reasoning scheme might consist of considering most probable states within a relevant subset of the network until the sum of the probabilities of the remaining states is below a small error threshold $\epsilon$. Incorporation of utility considerations into such algorithm and converting it into a normatively correct decision procedure is straightforward (Druzdzel, 1993). A crucial issue in such a computation is controlling for precision so that atypical symmetric models lead merely to loss of efficiency but not to incorrect posterior beliefs or inferior decisions. This can be accomplished by fitting a lognormal distribution to the joint probability distribution and computing the expectations based on this distribution, as shown in Section 3.2.

## Acknowledgments

While I am solely responsible for any possible remaining errors, discussions with several individuals improved the above argument significantly. Feedback obtained during my presentation of this work in CMU's Department of Philosophy allowed me to notice a major flaw in an early version of the paper. Thanks to Clark Glymour, Richard Scheines, Peter Spirtes, Greg Cooper, and others for their friendly criticism. Comments from Herb Simon helped in framing the argument. Anonymous reviewers provided useful feedback about the presentation.

---

[4](Druzdzel & Suermondt, 1994) review a variety of methods for determining relevance in the context of BBNs.